\documentclass[letterpaper, 10 pt, conference]{ieeeconf}  

\IEEEoverridecommandlockouts                              

\usepackage{graphicx} 
\usepackage{amsmath} 
\usepackage{amssymb}  
\usepackage[ruled,linesnumbered]{algorithm2e}
\usepackage{caption}
\usepackage{hyperref}
\hypersetup{hidelinks}
\usepackage{svg}
\title{\LARGE \bf
A Multi-Robot Task Assignment Framework for Search and Rescue with Heterogeneous Teams
}

\author{Hamid Osooli$^{1}$, Paul Robinette$^{2}$, Kshitij Jerath$^{3}$, and  S. Reza Ahmadzadeh$^{1}$
\thanks{$^{1}$ PeARL Lab, Richard Miner School of Computer and Information Sciences, University of Massachusetts Lowell, MA, USA {\tt\small \{hamid\_osooli, reza\_ahmadzadeh\}@uml.edu}}%
\thanks{$^{2}$ Department of Electrical and Computer Engineering, University of Massachusetts Lowell, MA, USA {\tt\small paul\_robinette@uml.edu}}%
\thanks{$^{3}$ Department of Mechanical Engineering, University of Massachusetts Lowell, MA, USA {\tt\small kshitij\_jerath@uml.edu}}
}%

\begin{document}
\maketitle
\thispagestyle{empty}
\pagestyle{empty}
\begin{abstract}
In post-disaster scenarios, efficient search and rescue operations involve collaborative efforts between robots and humans. Existing planning approaches focus on specific aspects but overlook crucial elements like information gathering, task assignment, and planning. Furthermore, previous methods considering robot capabilities and victim requirements suffer from time complexity due to repetitive planning steps. To overcome these challenges, we introduce a comprehensive framework\textemdash the Multi-Stage Multi-Robot Task Assignment. This framework integrates scouting, task assignment, and path-planning stages, optimizing task allocation based on robot capabilities, victim requirements, and past robot performance. Our iterative approach ensures objective fulfillment within problem constraints. Evaluation across four maps, comparing with a state-of-the-art baseline, demonstrates our algorithm's superiority with a remarkable 97 percent performance increase. Our code is open-sourced to enable result replication.
\end{abstract}

\section{Introduction}

Advancements in robotics enable their assistance in dangerous human-avoidant search and rescue tasks~\cite{murphy2001sar}. Such operations often demand coordination of multiple robots and humans, but this poses various challenges~\cite{queralta2020collaborative, khamis2015multi}. These challenges encompass diverse tasks in varied environments with constraints~\cite{queralta2020collaborative, rodriguez2020wilderness, liu2016multirobot}, underscoring the importance of effective multi-robot task assignment frameworks~\cite{grayson2014search}. Existing frameworks lack comprehensive inclusion of information gathering, assignment, and planning~\cite{carreno2020task,luo2013distributed}, and optimizing them becomes more complex with increased robot diversity~\cite{luo2015distributed}.

To address these issues, we propose a scouting-integrated framework for search and rescue encompassing Multi-Stage Multi-Robot Task Assignment (MSMRTA) and path planning. Our framework optimizes assignment of diverse-capability robots to specific tasks. Task assignment employs optimization and market-based techniques~\cite{badreldin2013comparative}, enhancing practicality. Empirical evaluation across various map complexities demonstrates significant planning time reduction (97\%) compared to a baseline (MRGA)~\cite{queralta2020collaborative}, with a slight (61.2\%) increase as robot-victim instances grow. This paper uses off-the-shelf methods for the scouting and path-planning stages of the framework, but our proposed framework can be extended to incorporate other search and planning algorithms as well.

\section{Related Work}
As a humanitarian research domain \cite{casper2000issues}, the development of robot technology for search and rescue operations continues to be an important area of study. To enhance the efficiency of multi-robot search and rescue, Quattrini Li et al. \cite{quattrini2016semantically} introduced semantic cues, while Liu et al. \cite{liu2017learning} employed hierarchical reinforcement learning. To further improve the overall performance of the system, we integrate semantics related to the capabilities and requirements of robots via binary vectors.

Multi-Robot Task Assignment (MRTA) is vital for efficient search and rescue missions, categorized as Centralized methods \cite{zhao2015heuristic}, Decentralized methods \cite{mouradian2017coalition}, and Hybrid methods \cite{liu2016multirobot,hussein2014multi}. For instance, Zhao et al. \cite{zhao2015heuristic} employed centralized methods to locate survivors using heterogeneous robots, while our approach utilizes a team of scouts for accelerated search.

Decentralized methods, exemplified by Mouradian et al. \cite{mouradian2017coalition}, utilize multi-objective optimization with inter-robot communication. In contrast, our framework employs a communication-independent ant colony algorithm for efficient scouting.

Hybrid methods, like Liu et al. \cite{liu2016multirobot}, combine autonomy and human intervention, while a hybrid framework models tasks as multiple traveling salesman problems \cite{hussein2014multi}. Our unique approach employs the $A^*$ algorithm independently for diverse calculations.

Further classification of MRTA involves optimization-based \cite{zhao2015heuristic,mouradian2017coalition} and market-based \cite{hussein2014multi} approaches. Both have strengths and weaknesses \cite{badreldin2013comparative}. In our paper, we propose the multi-stage multi-robot task assignment (MSMRTA) algorithm, integrating both techniques for enhanced effectiveness.

\section{Problem Description}

Prior to allocating victim requirements to the rescue team, essential information including location and requirements must be gathered through a scouting session using a set of uniform robots. These robots provide initial insights into the rescue operation's aspects such as victim count, locations, and needs. The main problem addressed in this study is the task allocation for a team of heterogeneous robots in a search and rescue mission in a building. The proposed solution is applicable to both homogeneous and heterogeneous robot groups, accommodating missions with varying demands. The introduced Multi-Stage Multi-Robot Task Assignment (MSMRTA) framework combines optimization-driven and market-based planning phases. A key aspect is the requirement analysis outlined in Algorithm~\ref{alg:ReqmentAnalysis}, within which the MSMRTA, including the $A^*$ path-planning algorithm for navigating building obstacles, allocates victim requirements to suitably capable robots. The framework's central process is illustrated in Algorithm~\ref{alg:mrta}, and assignment outcomes are stored in designated variables ($B2, B3, B4$). Reproducibility of results is facilitated through the open-source availability of the implementation code\footnote{Github repository: \url{https://github.com/hamidosooli/Multi-Robot-Task-Assignment}}.

\begin{algorithm}[ht]
\text{Input: map, $R$ (List of the Robots)}\\ 
\text{Output: Prioritized Path to each Victim}\\
$\text{V}\gets \texttt{Scouting}$(map)\hspace{1.5em}// Victims' information\\\label{Scouting}
$L^\text{full}, L^\text{partial}, L^\text{potential}\gets \texttt{ReqmentAnalysis}(R, V)$\\\label{ReqmentAnalysis}
$L^\text{unavailable}\gets \texttt{MissingCap}(V, L^\text{potential})$\\\label{MissingCap}
$C, C^\text{center}\gets \texttt{Clustering}(K, V)$\hspace{1.5em}// K-means\\\label{Clustering}
$B2, \overline{V}\gets \texttt{ClstrAsgn}(R, V, L^\text{full}, L^\text{partial}, C, C^\text{center})$\\\label{ClstrAsgn}
$T^\text{success}, T^\text{failure}\gets \texttt{PerfAnalysis}(R, \overline{V}, L^\text{partial})$\\\label{PerfAnalysis}
$B3, \overline{V}\gets \texttt{VictimAssign}(R, V, \overline{V}, T^\text{success}, T^\text{failure})$\\\label{VictimAssign}
$B4\gets \texttt{RobotAssign}(R, V, \overline{V}, L^\text{potential})$\\\label{RobotAssign}
\texttt{Path-Planning}($B2, B3, B4$)\hspace{1.5em}// $A^*$\label{Path-Planning}
\caption{Search and Rescue framework including Scouting, Multi Robot Task Assignment Analysis, Optimization, and Bidding stages, and Path-Planning}
\label{alg:mrta}
\end{algorithm}
\subsection{Scouting} 

\begin{figure}[ht]
\centering
\includegraphics[trim=5 5 5 5, clip, width=\linewidth]{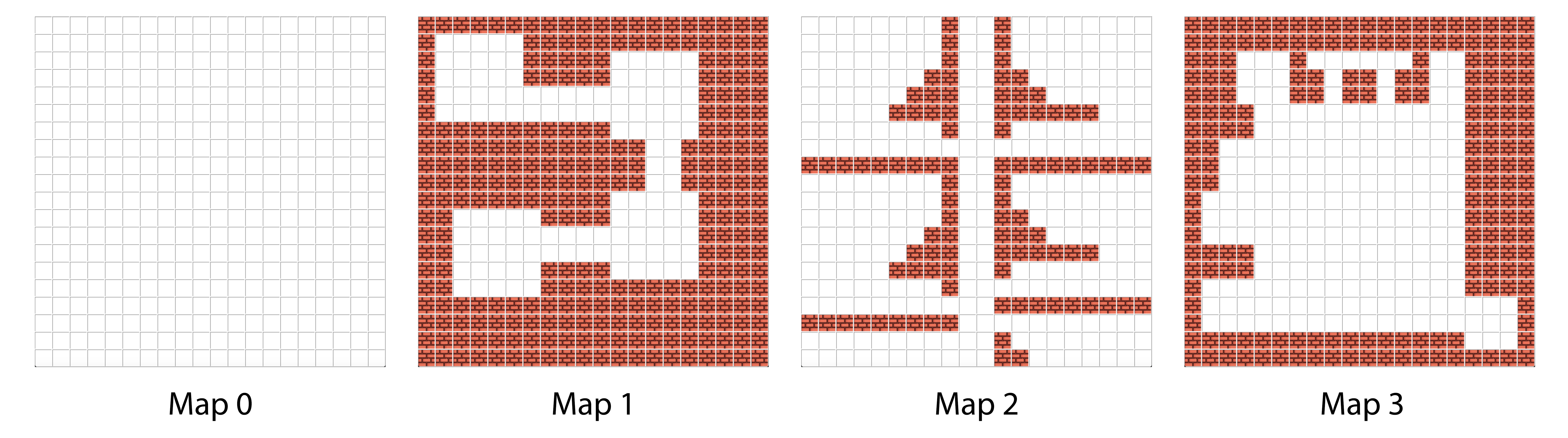}
  \caption{\small{Four $20\times 20$ maps used in our experiments. Obstacles have been depicted in red.}}
    \label{fig:maps}
\end{figure}

In post-disaster emergencies, promptly assessing the environment provides vital victim distribution data~\cite{nazarova2020application}. We employ a 2D gridworld simulation to model building layouts and attributes (Fig.\ref{fig:maps}). 

Our \texttt{Scouting} function (Algorithm\ref{alg:mrta}) uses the environment map to yield victim information—ID, location, requirements. Robots apply the ant colony grid search algorithm~\cite{dorigo1996ant}, mimicking ant search strategies and preventing redundant coverage via a search table. Scouts possess adaptable visual field depth (VFD) for victim detection, adjusting to obstacles. A scout identifies victims within its VFD vicinity (Fig.~\ref{fig:mission}), extracting details. Agents navigate by moving forward, backward, right, or left in the grid.

\subsection{Task Assignment} 

Inspired by the work in~\cite{carreno2020task}, we address the task assignment problem in the context of a heterogeneous robot fleet. Throughout the paper, we use the terms 'tasks' and 'victims' interchangeably to refer to the same entity. The robots can either completely fulfill the victim's requirements or fulfill part of the requirements, depending on their sensing and actuating capabilities. In line with the arguments presented in~\cite{badreldin2013comparative}, we propose a hybrid multi-stage task assignment framework integrating optimization and market-based methods. The proposed framework follows the taxonomy in~\cite{gerkey2004formal} and finds solutions for a group of single robot tasks (SR) through time-extended assignment (TA) of the tasks to multi task (MT) robots.
To better describe our system, we have created a list of capabilities/requirements for both the robots and victims. For the robots, we represent these capabilities as a binary vector, indicating which capabilities each robot possesses. Similarly, we have created a binary vector for the victims to represent their individual requirements ($\Vec{q}$ in Table~\ref{table:victims} and $\Vec{p}$ in Table~\ref{table:robots}).

\subsubsection{Victim Requirement and Robot Capability Analysis}
The \texttt{ReqmentAnalysis }(Algorithm~\ref{alg:ReqmentAnalysis} and line~\ref{ReqmentAnalysis} in Algorithm~\ref{alg:mrta}) focuses on determining three key sets of information through calculations. These sets include: victims that their requirements can be fully met by a given robot~($L^\text{full}$), victims that their requirements can be partially fulfilled by a given robot~($L^\text{partial}$), and potential robots that can meet a victim's requirements~($L^\text{potential}$).
\begin{algorithm}[ht]
\text{Input: $P$, $Q$}\\ 
\text{Output: $L^\text{full}, L^\text{partial}, L^\text{potential}$}\\

$U\gets Q P^\top$// Matrix Multiplication\\\label{matmul}
$S\gets \sum_{j=0}^{nr-1} Q[:, j]$\hspace{1.5em}// nr = number of requirements\\
$L^\text{full}\gets (U > 0)\hspace{.5em} and \hspace{.5em}(U == S)$\\
$L^\text{partial}\gets (U > 0)\hspace{.5em} and \hspace{.5em}(U < S)$\\
\For{i in $(L^\text{full}\hspace{.5em} and \hspace{.5em}L^\text{partial})$}{
$L^\text{potential}\gets Q[i(0), :]\hspace{.5em} and \hspace{.5em}Q[i(1), :]$\\
}

\caption{The $\texttt{ReqmentAnalysis}$ function identifies victims fully satisfied ($L^\text{full}$), partially satisfied ($L^\text{partial}$) by a robot, and potential robots meeting a victim's requirements ($L^\text{potential}$).}
\label{alg:ReqmentAnalysis}
\end{algorithm}
When a robot fulfills all victim requirements, the victim ID is included in the set $L^\text{full}$. If a robot satisfies only a subset of victim requirements, the corresponding victim ID is added to the robot's $L^\text{partial}$ set. Conversely, robots capable of meeting a victim requirement contribute their IDs to the $L^\text{potential}$ set for that requirement. By representing victim requirements ($\Vec{q}$) and robot capabilities ($\Vec{p}$) in matrix form, the system performs a matrix operation (Algorithm~\ref{alg:ReqmentAnalysis}, line~\ref{matmul}) to compare the outcome with a larger vector encompassing all requirements (S) (Exemplified in Tables~\ref{table:victims} and~\ref{table:robots}). These binary vectors signify agent possession of each attribute. The algorithm's inputs are matrices $Q = [\vec{q_0}^\top, \ldots, \vec{q_{M-1}}^\top]^\top$ and $P = [\vec{p_0}, \ldots, \vec{p_{N-1}}]$ containing information, presented as matrices to mitigate time complexity.

Using the data in $L^\text{potential}$, the \texttt{MissingCap} function (Algorithm~\ref{alg:mrta}, line~\ref{MissingCap}) identifies instances where no suitable robots exist to fulfill specific requirements. Subsequently, it appends the relevant victim IDs and requirements to $L^\text{unavailable}$, accounting for deficient capabilities.

\subsubsection{Clustering of Victims and Allocation of Victim Clusters to Robots}

In order to facilitate the task assignment process, victims are grouped into clusters based on their map locations using the \texttt{Clustering} function, employing the K-means algorithm~\cite{macquuen1967some}. These clusters, equal in number to the robots~($K~=~N$), enable efficient robot-to-victim group assignments for proximate victims.

Robot-cluster matching employs a linear assignment approach~\cite{luo2013distributed}, our assignment formulation(Equation~\ref{eq:clusterAssignment}) incorporates two assignment weights: $w_{rc}^f$ (denoting victim requirement sets fully satisfied by robot $r$ in cluster $c$) and $w_{rc}$ (indicating total victim requirements met by robot $r$ in cluster $c$) for $N$ robots and $K$ clusters.

\begin{equation}
     \begin{aligned}
     &\underset{\{x_{rc}\}}{\text{max }} \sum_{r=0}^{N-1}\sum_{c=0}^{K-1} [(1-\psi) w_{rc} + \psi w_{rc}^f]x_{rc}\\&
     \text{s.t. }\sum_{r=0}^{N-1} x_{rc} \leq 1,\quad \forall c = 0, 1, ..., K-1\\&
     x_{rc} \in \{0, 1\},\quad \forall r, c
     \end{aligned}
    \label{eq:clusterAssignment}
\end{equation}

The importance coefficient $\psi\in \left[0, 1\right]$ balances fulfilling victim's requirements fully versus partially. A high $\psi$ prioritizes robots in clusters satisfying one or more victims fully, while a low $\psi$ favors clusters partially meeting requirements of more victims. Binary variable $x_{rc}$ decides robot assignment to clusters. The equation's second line ensures a cluster is assigned to at most one robot~\cite{luo2013distributed}.

Once robots join clusters, they move to cluster centers, calculated as victims' location averages. Manhattan distance from center to victims guides prioritization within clusters. The \texttt{ClstrAsgn} function (Algorithm~\ref{alg:mrta} line~\ref{ClstrAsgn}) computes this, storing result in $B2$.

\begin{figure}[ht]
  \centering
  \includegraphics[trim=0 110 0 120, clip, width=\columnwidth]{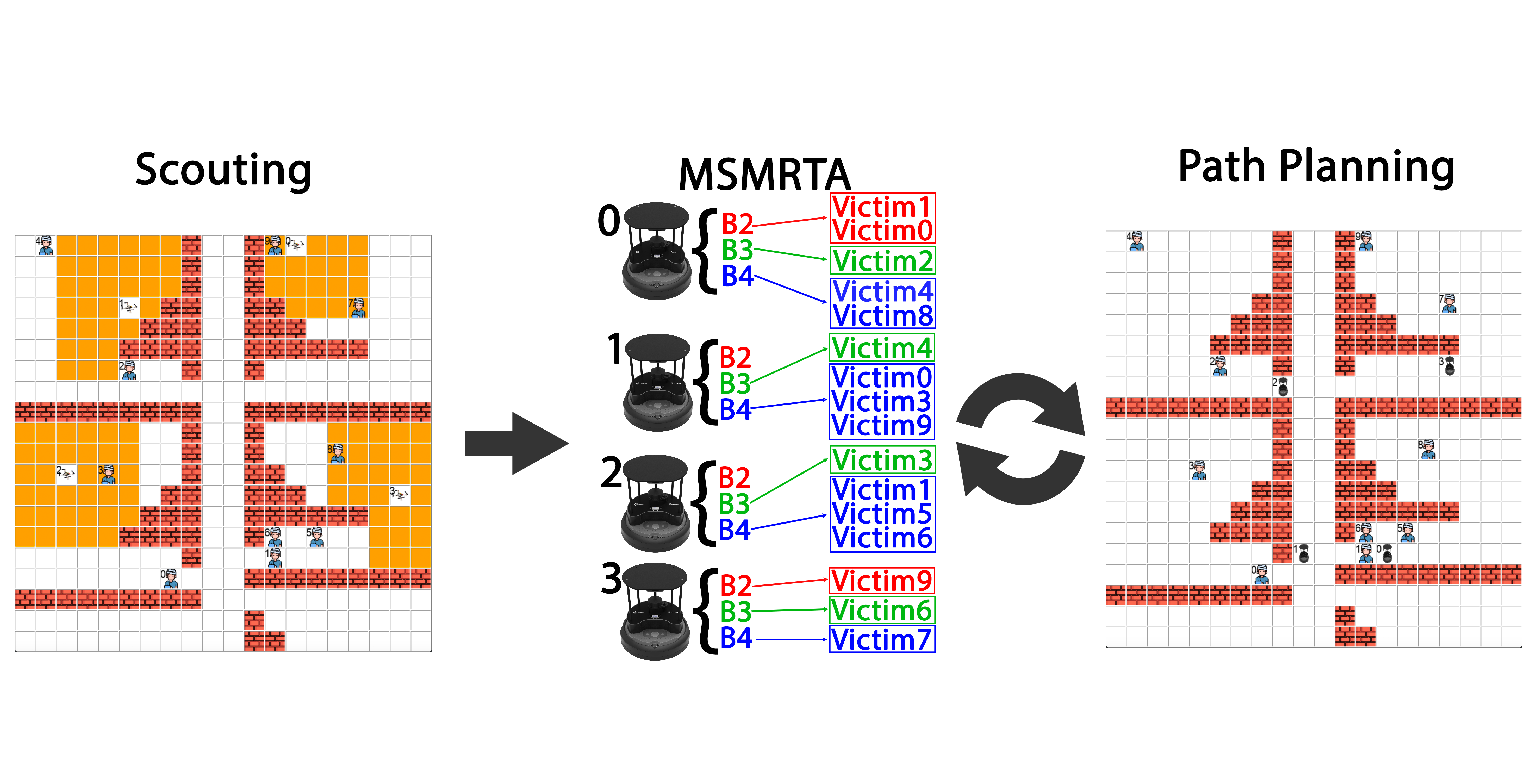}
  \caption{\small{Various steps of our proposed framework. The locations of victims and scouts (left). Each scout's Visual Field Depth (VFD) has been plotted in orange.
  The output of the different assignment stages (middle). The path planning step (right).}}
  \label{fig:mission}
\end{figure}

A clustering method that relies solely on location (\cite{carreno2020task,janati2017multi}) can be insufficient because it may fail to account for obstacles in the environment. For example, victims on opposite sides of a wall may be grouped together in the same cluster, leading to suboptimal robot travel times. To overcome this issue, we devised a method that examines each cluster for obstacles by comparing the Manhattan distance from the cluster center to each victim with that of an empty environment. If the distance in the empty environment is greater than the distance in the environment with obstacles (walls), it suggests the existence of an obstacle, and the victim is excluded until the subsequent stage.

\subsubsection{Analysis of Robot Performance and Reliability}

Robots' performance in assigned clusters involves individual interactions with victims, either succeeding or failing. Evaluating reliability relies on analyzing time allocation for successful ($T^\text{success}_{rv}$) and unsuccessful ($T^\text{failure}_{rv}$) tasks. Robot reliability is computed to minimize both time categories, enhancing overall reliability. This optimization employs linear assignment to manage busy time for $N$ robots and $M$ victims using Equation~\ref{eq:spentTime}. Binary variable $x_{rv}$ ensures exclusive victim-to-robot assignment~\cite{luo2013distributed}.

A coefficient $\beta\in \left[0, 1\right]$ balances the time cost trade-off. Success and failure times are calculated using \texttt{PerfAnalysis} function (Algorithm~\ref{alg:mrta} line~\ref{PerfAnalysis}). Subsequently, \texttt{VictimAssign} (Algorithm~\ref{alg:mrta} line~\ref{VictimAssign}) employs this data to assign remaining victims to robots, stored as variable $B3$.

\begin{equation}
     \begin{aligned}
     &\underset{\{x_{rv}\}}{\text{min }}\sum_{v=0}^{M-1}\sum_{r=0}^{N-1}[(1-\beta) T^\text{success}_{rv} + \beta T^\text{failure}_{rv}]x_{rv}\\&     
     \text{s.t. }\sum_{r=0}^{N-1} x_{rv} \leq 1,\quad \forall v = 0, 1, ..., M-1\\&
     x_{rv} \in \{0, 1\},\quad \forall r, v
     \end{aligned}
\label{eq:spentTime}
\end{equation}

\subsubsection{Bidding Based on Distance}

We employ the potential robot list ($L^\text{potential}$) from the \texttt{ReqmentAnalysis} function to match unattended victim requirements with the nearest robot. The proximity is determined using Manhattan distance computation among potential robots. This approach employs a sealed bid scheme wherein robots submit concealed bids until auction culmination. The winner is the highest bidder, which in this context is the closest robot to the victim. The outcome, denoted as $B4$, is computed via the \texttt{RobotAssign} function (Algorithm~\ref{alg:mrta}, line~\ref{RobotAssign}), ensuring comprehensive fulfillment of victim requirements. Additionally, we verify $L^\text{full}$ to prevent duplication. Specifically, if a robot is capable of fulfilling all of a victim's requirements, then that victim should not be assigned to any other robots.

\subsection{Path-Planning} 
The~\texttt{Path-Planning} function in Algorithm~\ref{alg:mrta} line~\ref{Path-Planning} uses the $A^*$ search algorithm, that is a widely used path-planning algorithm and particularly well suited for grid-based environments. $A^*$ calculates the shortest path from the robot's initial location to the assigned victim's address by taking into account both the distance traveled and the estimated distance to the goal, also known as the heuristic cost. In our system, we use the Manhattan distance as the heuristic cost. The $A^*$ algorithm ensures that the robot's path is efficient, while also avoiding obstacles and ensuring safe navigation. The calculated paths are then used by the robots to navigate to the assigned victims.
\begin{figure}[h]
\centering
\includegraphics[trim=3em 4em 10em 5em, clip, width=0.49\linewidth]{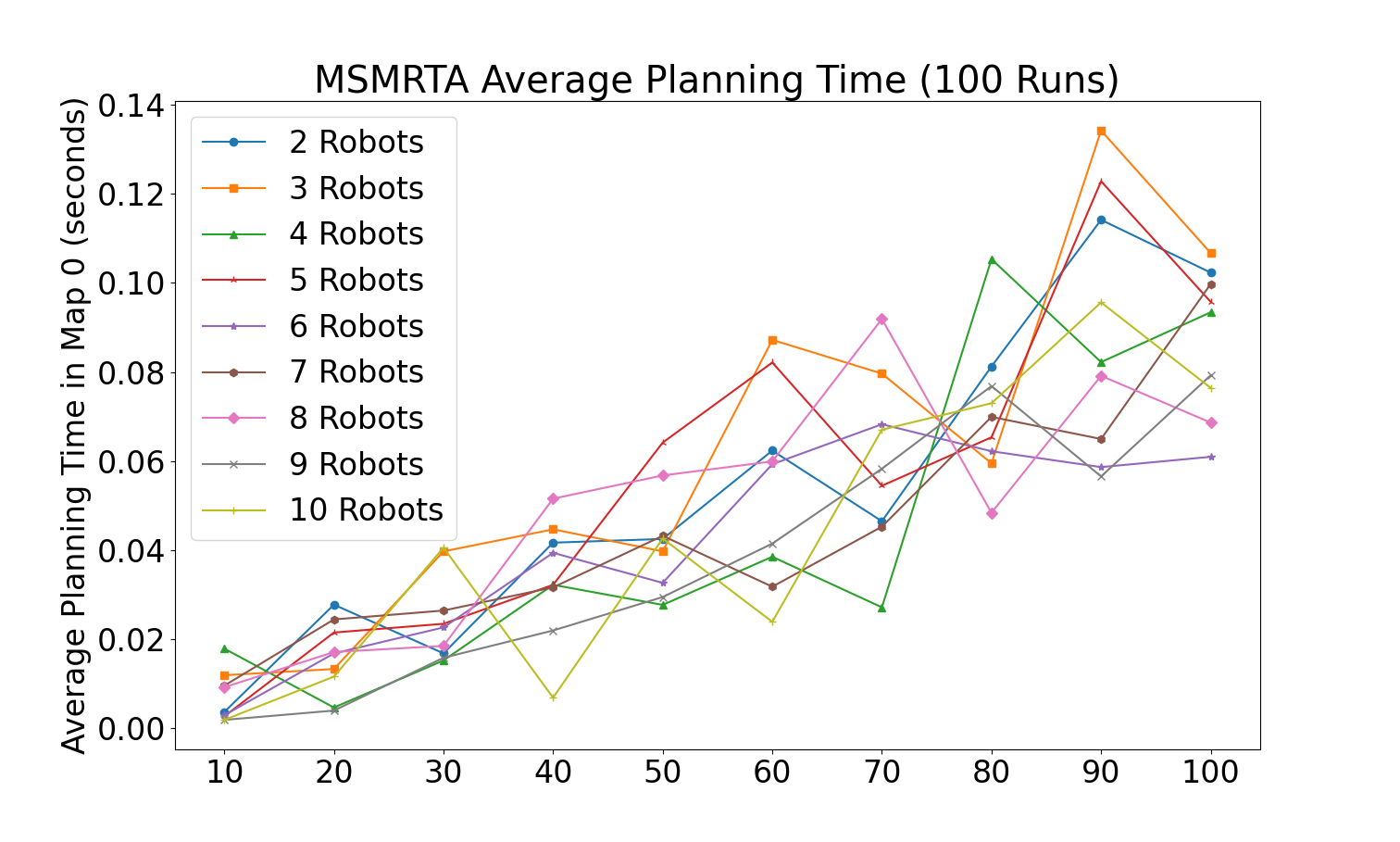}
\includegraphics[trim=5em 4em 10em 5em, clip, width=0.49\linewidth]{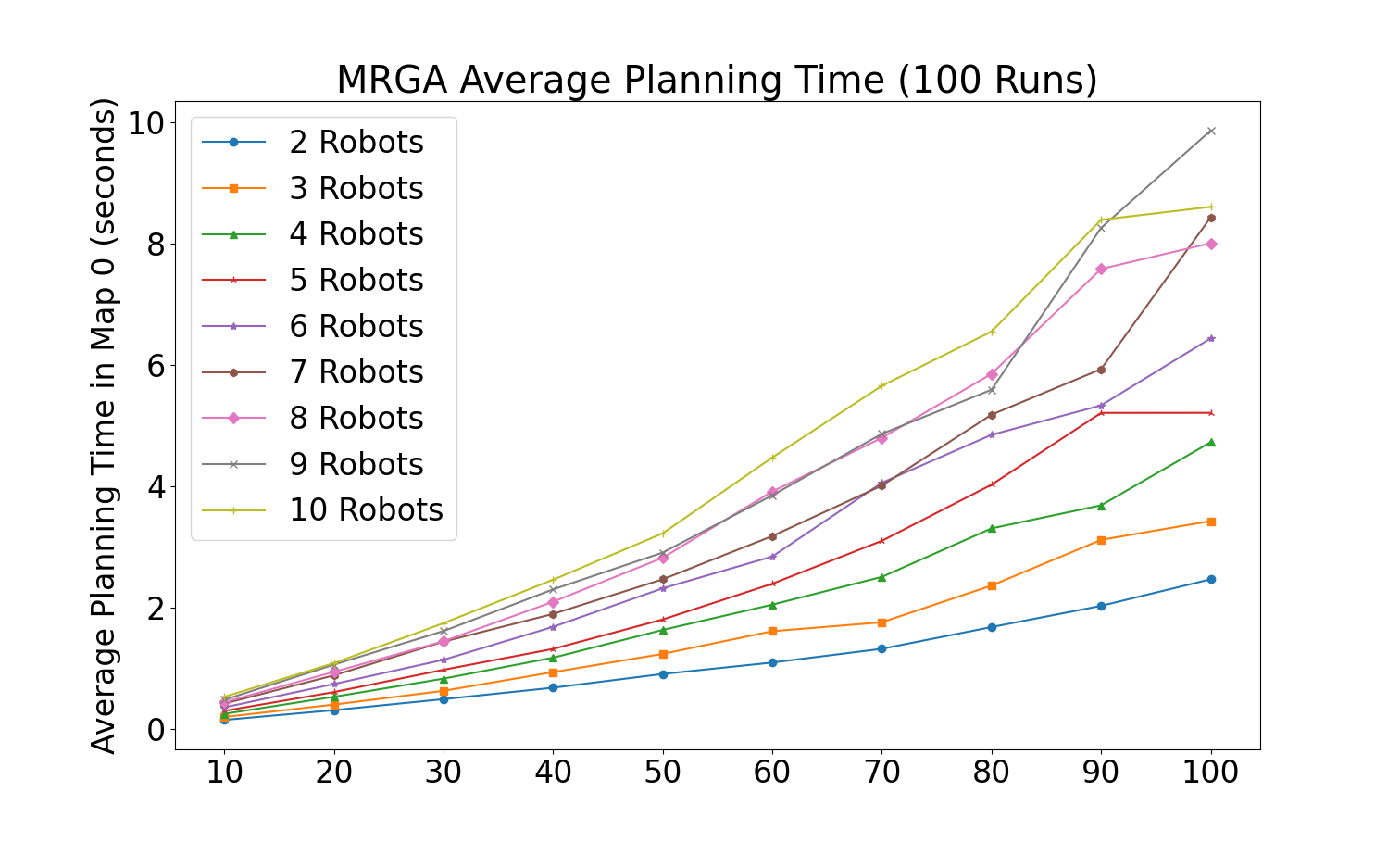}
\includegraphics[trim=3em 4em 10em 5em, clip, width=0.49\linewidth]{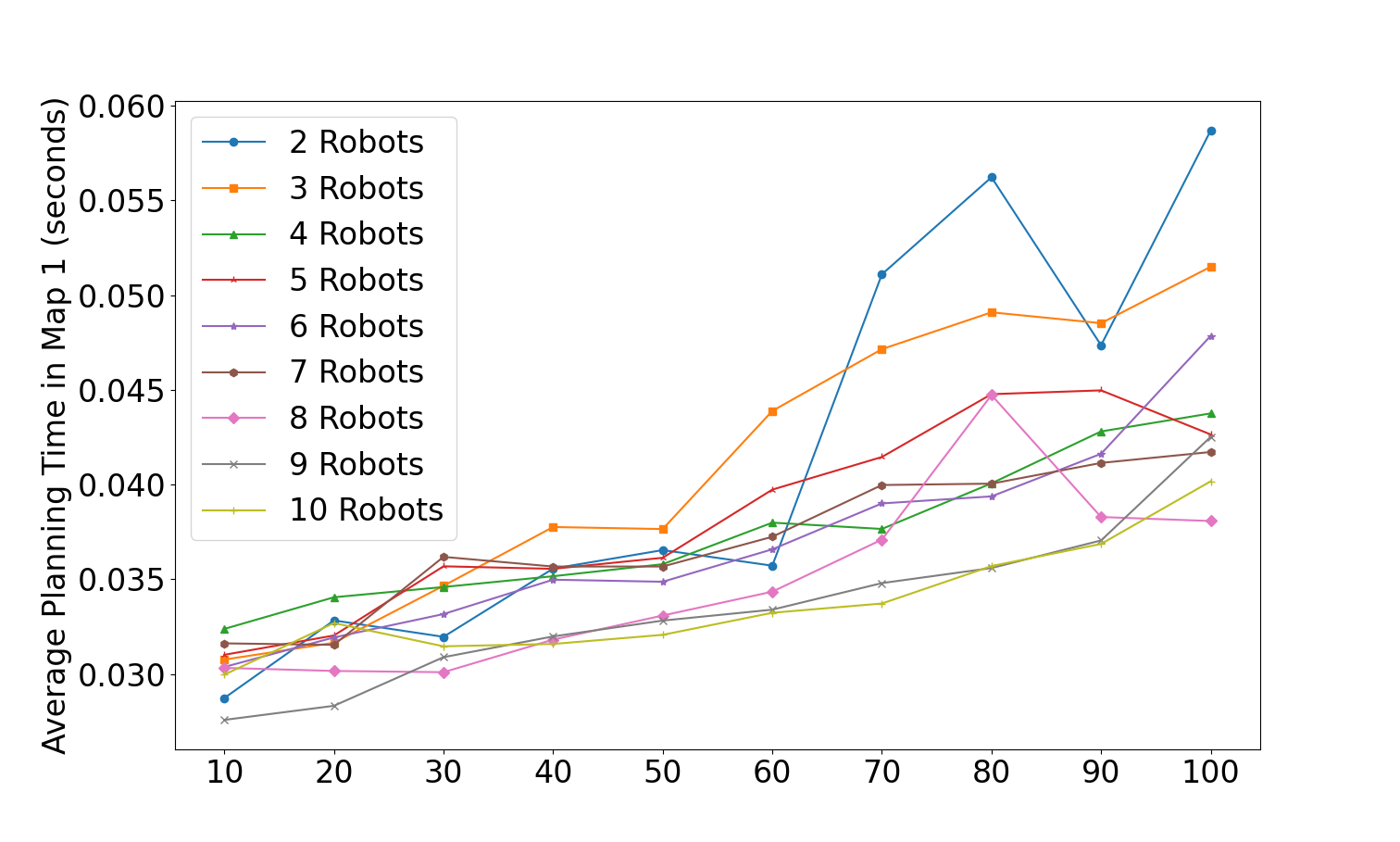}
\includegraphics[trim=5em 4em 10em 5em, clip, width=0.49\linewidth]{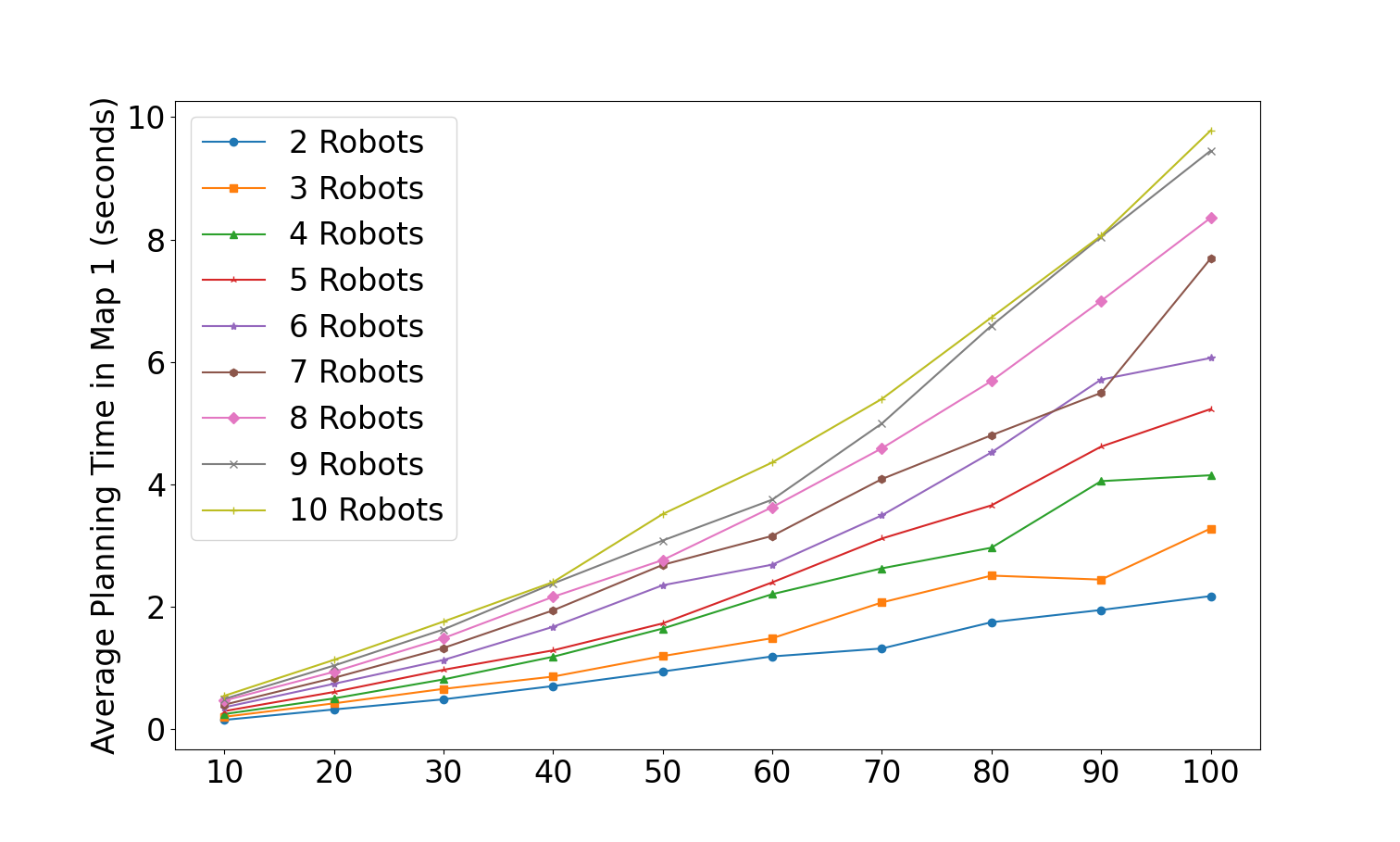}

\includegraphics[trim=3em 4em 10em 5em, clip, width=0.49\linewidth]{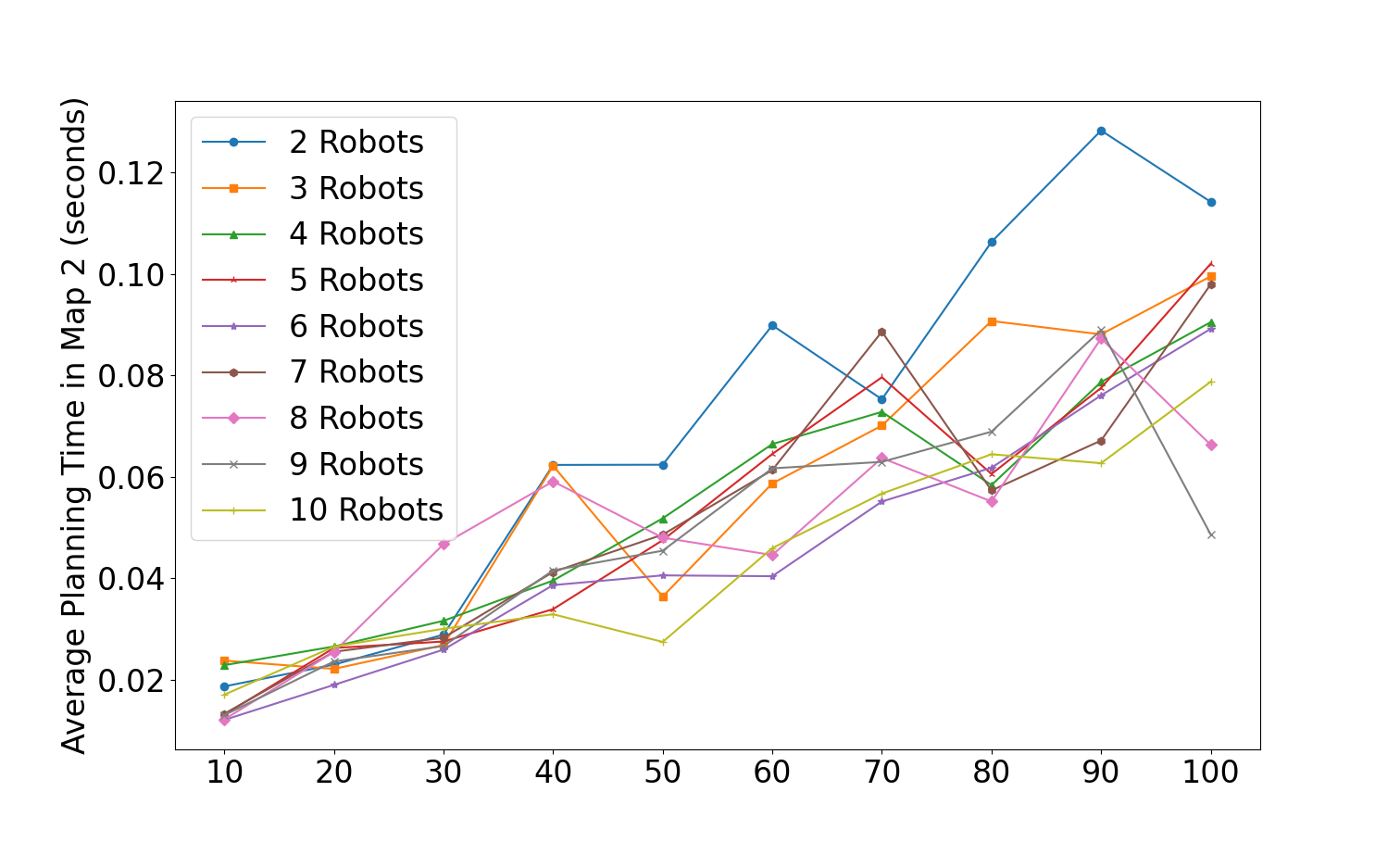}
\includegraphics[trim=5em 4em 10em 5em, clip, width=0.49\linewidth]{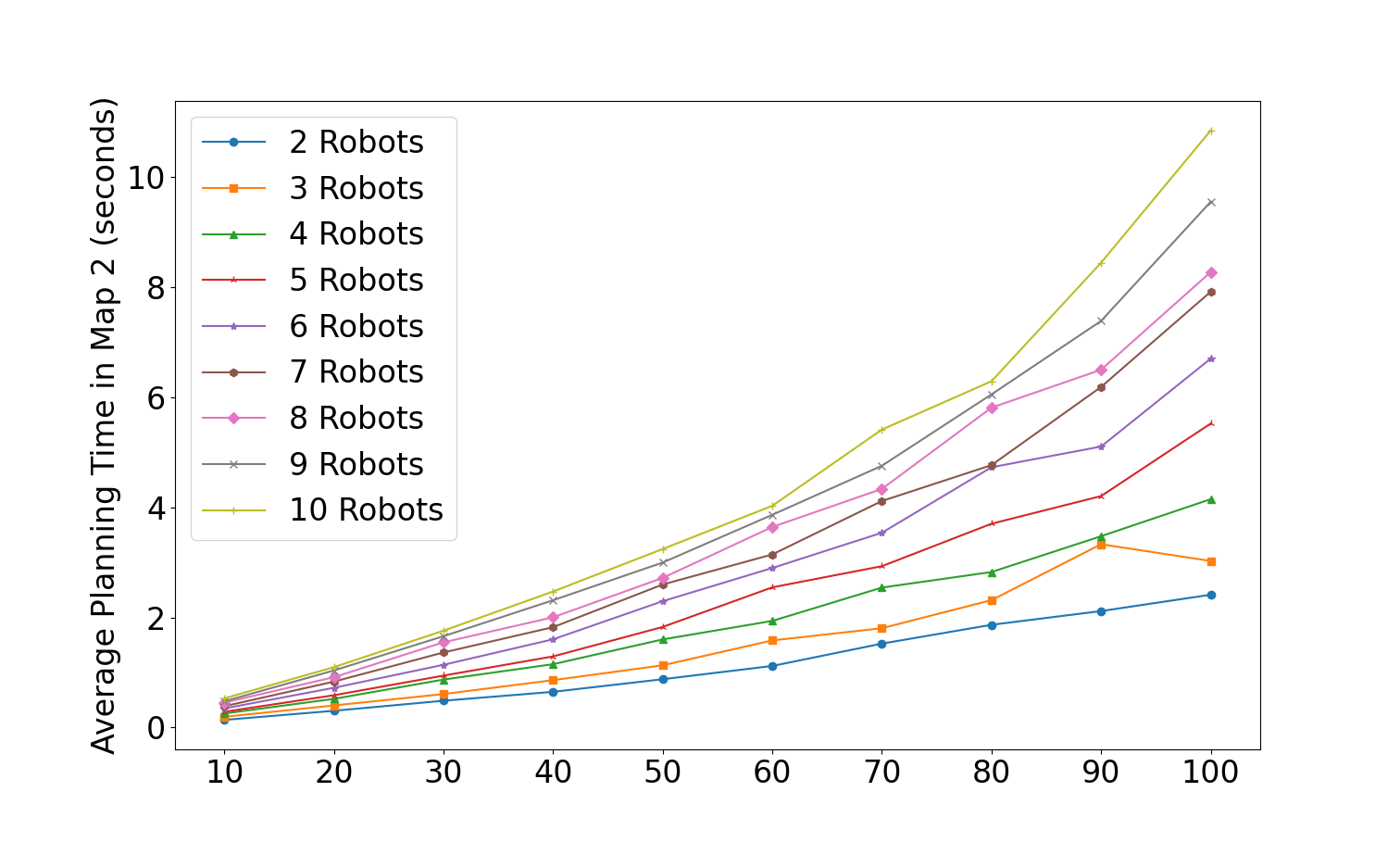}
\includegraphics[trim=3em 1em 10em 5em, clip, width=0.49\linewidth]{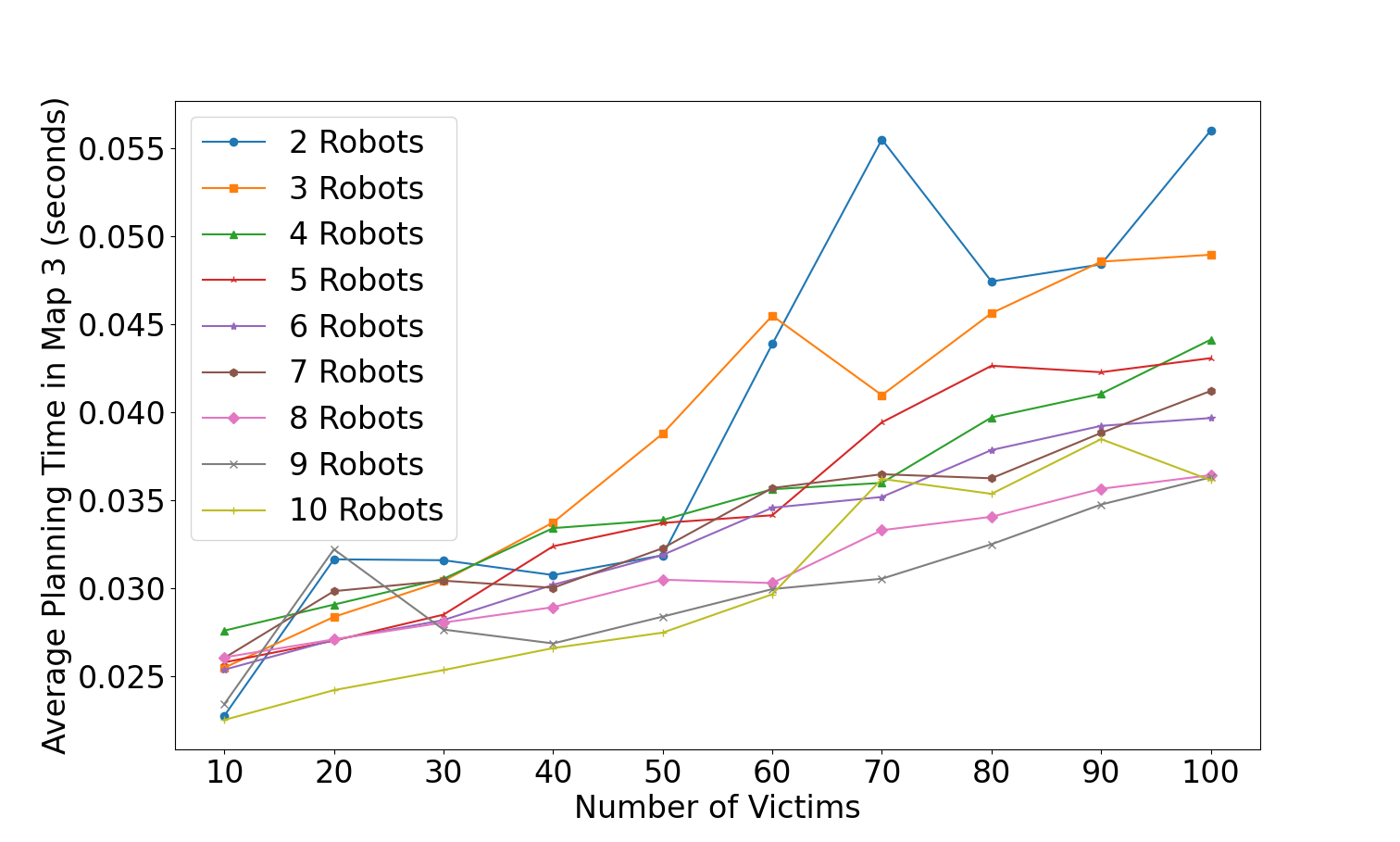}
\includegraphics[trim=5em 1em 10em 5em, clip, width=0.49\linewidth]{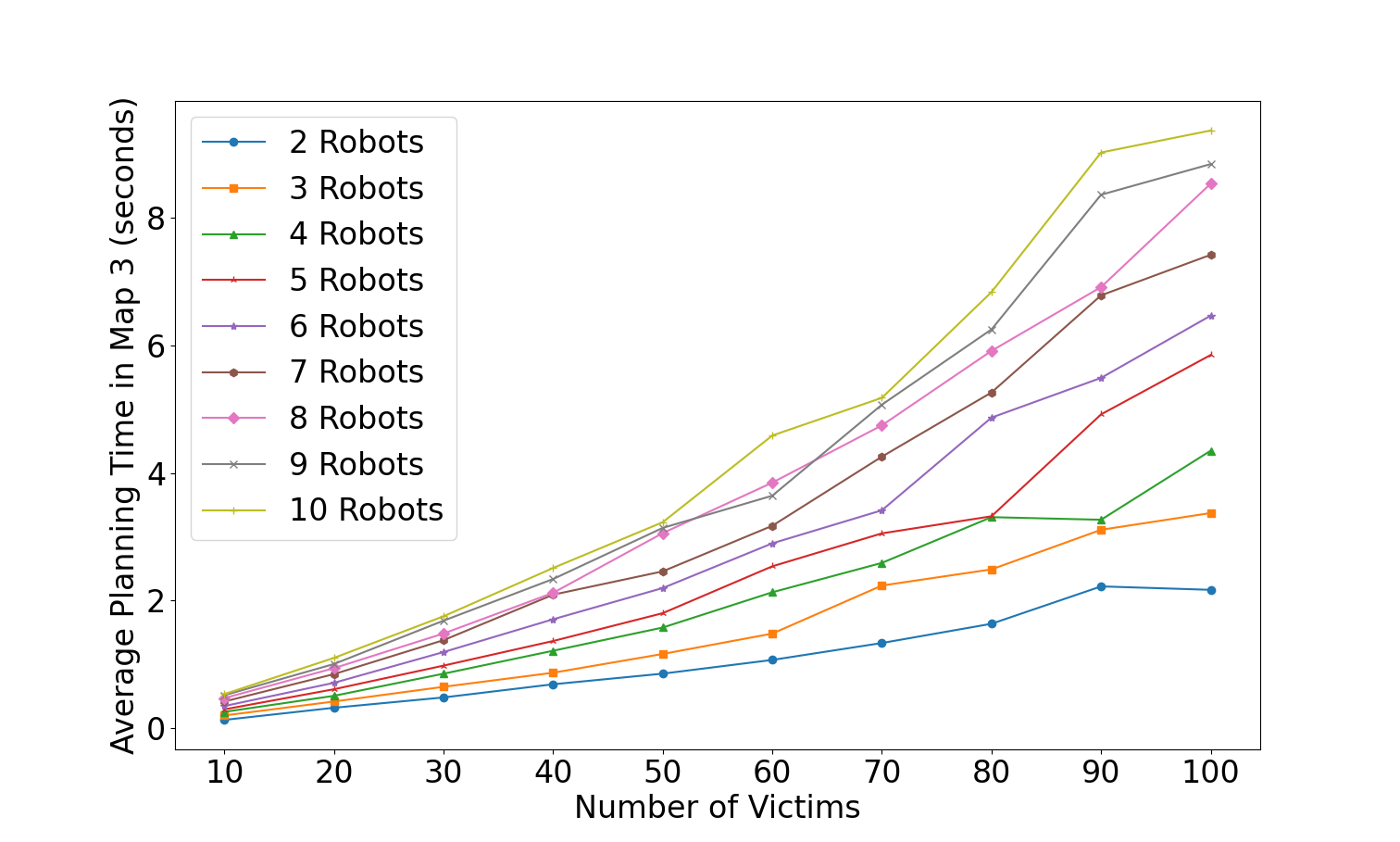}
	\caption{\small{Average planning time for the proposed algorithm (MSMRTA) on the left and the baseline algorithm (MRGA) on the right. 
 }}
	\label{fig:comparison}
\end{figure}
\section{Results \& Discussion}
In this section, we present the experimental results of our proposed framework on \textit{Map 2}. Ten victims with diverse requirements (Table~\ref{table:victims}) were assigned to four robots with varying capabilities (Table~\ref{table:robots}). The Ant colony algorithm was employed by the scouting team for environment coverage before assignment. All ten victims were successfully located, and pertinent data such as ID, location, and requirements were gathered and recorded for task allocation (Table~\ref{table:victims}). The Cluster ID in Table~\ref{table:victims} indicates clustering outcomes (Algorithm~\ref{alg:mrta}, line~\ref{Clustering}), while ${L^\text{potential}}$ denotes a requirement/capability analysis output (Algorithm~\ref{alg:mrta}, line~\ref{ReqmentAnalysis}). The remaining columns exhibit scouting team-acquired victim information.
\begin{table}[ht]
	\caption{\small{Summary of victims' locations, requirements, and potential robots for fulfillment of the requirements, as well as the corresponding cluster ID for each victim}}
 \label{table:victims}
	\centering
\setlength\tabcolsep{1pt}
 \begin{tabular}{ c c c c c }
		Victim ID & Location & Cluster & $\Vec{q}$ & ${L^\text{potential}}$\\
		\hline
        0 & (16, 7) & $C_0$ & $[0, 1, 1, 0, 0, 0]^\top$ & $[\emptyset, [0], [1], \emptyset, \emptyset, \emptyset]$\\
        1 & (15, 12) & $C_0$ & $[0, 0, 0, 1, 1, 0]^\top$ & $[\emptyset, \emptyset, \emptyset, [0, 3], [1, 2], \emptyset]$\\
        2 & (6, 5) & $C_2$ & $[0, 1, 0, 0, 0, 1]^\top$ & $[\emptyset, [0], \emptyset, \emptyset, \emptyset, \emptyset]$\\
        3 & (11, 4) & $C_2$ & $[0, 0, 1, 0, 1, 0]^\top$ & $[\emptyset, \emptyset, [1], \emptyset, [1, 2], \emptyset, ]$\\
        4 & (0, 1) & $C_1$ & $[1, 1, 0, 0, 0, 0]^\top$ & $[[1, 3], [0], \emptyset, \emptyset, \emptyset, \emptyset]$\\
        5 & (14, 14) & $C_0$ & $[0, 0, 0, 0, 1, 1]^\top$ & $[\emptyset, \emptyset, \emptyset, \emptyset, [1, 2], \emptyset]$\\
        6 & (14, 12) & $C_0$ & $[1, 0, 0, 0, 1, 0]^\top$ & $[[1, 3], \emptyset, \emptyset, \emptyset, [1, 2], \emptyset]$\\
        7 & (3, 16) & $C_3$ & $[1, 0, 0, 1, 0, 0]^\top$ & $[[1, 3], \emptyset, \emptyset, [0, 3], \emptyset, \emptyset]$\\
        8 & (10, 15) & $C_0$ & $[0, 1, 0, 1, 0, 0]^\top$ & $[\emptyset, [0], \emptyset, [0, 3], \emptyset, \emptyset]$\\
        9 & (0, 12) & $C_3$ & $[1, 0, 0, 0, 1, 0]^\top$ & $[[1, 3], \emptyset, \emptyset, \emptyset, [1, 2], \emptyset]$\\
	\end{tabular}
\end{table}

This representation indicates that victim~0 has requirements that include the second and third items of the capability/requirement list, and robot~0 and robot~1 are potential candidates to fulfill these requirements, respectively. Additionally, victim~0 is located at position (16, 7), which belongs to cluster $C_0$.

\begin{table}[ht]
	\caption{\small{Rescue Robots' locations and capabilities. List of the victims IDs whose requirements can be fully or partially met by each robot, as well as the result of the victims IDs assigned to each robot in each stage of the task assignment process.}}
 \label{table:robots}
	\centering
 \setlength\tabcolsep{1.5pt}
	\begin{tabular}{ c c c c c c }
		Robot ID & Location & $\Vec{p}$ & ${L^\text{full}}$ & ${L^\text{partial}}$ & Cluster\\
		\hline
        $0$ & (0, 9) & $[0, 1, 0, 1, 0, 0]^\top$ & [8] & [0, 1, 2, 4, 7] & $C^\text{center}_0$\\
        $1$ & (0, 10) & $[1, 0, 1, 0, 1, 0]^\top$ & [3, 6, 9] & [0, 1, 4, 5, 7] & $\emptyset$\\
        $2$ & (19, 9) & $[0, 0, 0, 0, 1, 0]^\top$ & $\emptyset$ & [1, 3, 5, 6, 9] & $\emptyset$\\
        $3$ & (19, 10) & $[1, 0, 0, 1, 0, 0]^\top$ & [7] & [1, 4, 6, 8, 9] & $C^\text{center}_3$
	\end{tabular}
\end{table}

Table~\ref{table:robots} shows robot~0's capabilities and assignments. Victims 2 and 5 have unavailable requirements as indicated by $L^\text{unavailable}$. Fig.\ref{fig:mission} illustrates assignment stages with color-coded outcomes. Robot~2 lacks clusters in B2 but gains tasks in bidding. Initially, robot~0 is assigned to victim~8, later adjusted for spatial constraints. Redundancy in B4 is minimized, except for nearest qualified robot.

Our MSMRTA algorithm was evaluated against Carreno et al.'s MRGA on 2 to 10 robots, 10 to 100 victims, and 4 maps. MSMRTA reduces planning times significantly: $97.70\%$, $97.00\%$, $96.93\%$, and $97.34\%$. Planning time increased with more robots and victims: $92.26\%$, $31.80\%$, $80.85\%$, and $39.89\%$. Fig.~\ref{fig:comparison} depicts these outcomes, highlighting MSMRTA's potential in enhancing search and rescue efficiency.

\section{Conclusion \& Future Work}
This paper introduces a novel search and rescue framework integrating scouting, multi-stage multi-robot task assignment, and path-planning. The framework leverages robot capabilities and victim requirements during scouting, optimizing task assignment based on this data and robots' past performance. Notably, it accommodates multiple assignments by different robots to fulfill all requirements. The framework's effectiveness was validated through simulations, exhibiting a 97\% average reduction in planning time and overall performance enhancement. Future work entails refining clustering techniques, analyzing environment maps, exploring influencing factors, and validating the algorithm via real-world experiments with diverse robotic fleets.






\section{Acknowledgments}

This work is supported in part by NSF (IIS-2112633) and the Army Research Lab (W911NF20-2-0089).

\bibliographystyle{IEEEtran}
\bibliography{bibtex/bib/ref}

\end{document}